\documentclass[conference]{Format/IEEEtran}

\usepackage{acro}
\usepackage[utf8]{inputenc}
\usepackage[ruled,vlined,linesnumbered]{algorithm2e}
\usepackage{amsfonts, amsmath, amssymb}
\usepackage{array}
\usepackage{bm}
\usepackage{cite}
\usepackage{color}
\usepackage{float}
\usepackage{graphicx}
\usepackage{subfig}
\usepackage{hyperref}
\usepackage{url}
\usepackage{multirow}
\usepackage{booktabs}

\usepackage{tabularx}
\usepackage[ancient]{flushend} 

\usepackage{tikz}

\usepackage{xcolor}
\colorlet{secondary}{white!25!black}

\usetikzlibrary{mindmap,trees,fit,decorations.pathreplacing,patterns,decorations.markings}
\newcounter{mynode}
\tikzset{step node/.code={\stepcounter{mynode}
}}

\tikzstyle{pyschologyBar} = [rounded corners=1mm,fill=cyan, draw=white]
\tikzstyle{systemBBar} = [rounded corners=1mm,fill=blue!45!green, draw=white]
\tikzstyle{riseFPBar} = [rounded corners=1mm,fill=blue!40!white, draw=white]
\tikzstyle{risePPBar} = [rounded corners=1mm,fill=red!40!white, draw=white]
\tikzstyle{firstDBBar} = [rounded corners=1mm,fill=orange, draw=white]
\tikzstyle{firstWKBar} = [rounded corners=1mm,fill=green!70!red, draw=white]
\tikzstyle{riseEBar} = [rounded corners=1mm,fill=yellow!70!black, draw=white]
\tikzstyle{primitiveBar} = [rounded corners=1mm,fill=blue!60!red, draw=white]
\tikzstyle{updatingBar} = [rounded corners=1mm,fill=red!60!black, draw=white]
\tikzstyle{educationText} = [rectangle, above=.1cm, align=center,,scale=1.0]
\tikzstyle{experienceText} = [rectangle, below=.1cm, align=center,,scale=1.00]

\title{Underwater Robot Manipulation: Advances, Challenges and Prospective Ventures}

\author{
    Sara Aldhaheri, Giulia De Masi, {\`E}ric Pairet, Paola Ard{\'o}n\\
    Technology Innovation Institute (TII), Abu Dhabi, United Arab Emirates \\
    \{\textit{sara.aldhaheri, giulia.demasi, eric.pairet, paola.ardon}\}\!\atsign tii.ae
}

\newcommand\atsign{@}

\newcommand*{\sref}[1]{Section~\ref{#1}}

\newcommand*{\fref}[1]{Figure~\ref{#1}}

\definecolor{purple}{RGB}{106,13,173}

\DeclareAcronym{1D}{
  short = 1D,
  long  = one-dimensional
}
\DeclareAcronym{2D}{
  short = 2D,
  long  = two-dimensional
}
\DeclareAcronym{3D}{
  short = 3D,
  long  = three-dimensional
}
\DeclareAcronym{AI}{
  short = AI,
  long  = artificial intelligence
}
\DeclareAcronym{AMADEUS}{
  short = AMADEUS,
  long  = Advanced Manipulation for Deep Underwater Sampling 
}
\DeclareAcronym{AUV}{
  short = AUV,
  long  = autonomous underwater vehicle
}
\DeclareAcronym{CAN}{
  short = CAN,
  long  = controller area network
}
\DeclareAcronym{DoF}{
  short = DoF,
  long  = degree of freedom,
  long-plural-form = degrees of freedom
}
\DeclareAcronym{HRI}{
  short = HRI,
  long  = human-robot interaction
}
\DeclareAcronym{I-AUV}{
  short = I-AUV,
  long  = Intervention autonomous underwater vehicle
}
\DeclareAcronym{INS}{
  short = INS,
  long  = inertial navigation system
}
\DeclareAcronym{LED}{
  short = LED,
  long  = light-emitting diode
}
\DeclareAcronym{LMS}{
  short = LMS,
  long  = least mean squares
}
\DeclareAcronym{NATO}{
  short = NATO,
  long  = North Atlantic Treaty Organization
}
\DeclareAcronym{NMSE}{
  short = NMSE,
  long  = normalised mean squared error
}
\DeclareAcronym{NN}{
  short = NN,
  long  = neural network
}
\DeclareAcronym{OBTLC}{
  short = OBTLC,
  long  = Object-Based Task-Level Control 
}
\DeclareAcronym{ODIN}{
  short = ODIN,
  long  = Omni-Directional Intelligent Navigator
}
\DeclareAcronym{OMPL}{
  short = OMPL,
  long  = Open Motion Planning Library
}
\DeclareAcronym{OTTER}{
  short = OTTER,
  long  = Ocean Technology Testbed for Engineering Research
}
\DeclareAcronym{PD}{
  short = PD,
  long  = proportional-derivative
}
\DeclareAcronym{ROS}{
  short = ROS,
  long  = robot operating system
}
\DeclareAcronym{ROV}{
  short = ROV,
  long  = remotely operated vehicle
}
\DeclareAcronym{SAUVIM}{
  short = SAUVIM,
  long  = Semi Autonomous Underwater Vehicle for Intervention Mission 
}
\DeclareAcronym{SAR}{
  short = SAR,
  long  = search and rescue
}
\DeclareAcronym{TP}{
  short = TP,
  long  = task priority
}
\DeclareAcronym{TWINBOT}{
  short = TWINBOT,
  long  = Twin Robots for Cooperative Underwater Intervention Missions
}
\DeclareAcronym{UNION}{
  short = UNION,
  long  = Underwater Intelligent Operation and Navigation 
}
\DeclareAcronym{UVMS}{
  short = UVMS,
  long  = underwater vehicle manipulator system
}
\begin{document}
    \maketitle

\begin{abstract}

    
    Underwater manipulation is one of the most remarkable ongoing research subjects in robotics. \acp{I-AUV} not only have to cope with the technical challenges associated with traditional manipulation tasks but do so while currents and waves perturb the stability of the vehicle, and low-light, turbid water conditions impede perceiving the surroundings. Certainly, the dynamic nature and our limited understanding of the marine environment hinder the autonomous performance of underwater robot manipulation. This manuscript provides a discussion on previous research and the limiting factors that impose on the long-envisioned prospects of autonomous underwater manipulation to finally highlight research directions that have the potential to improve the autonomy capabilities of \acp{I-AUV}.
\end{abstract}
    \section{INTRODUCTION}

Robotic manipulators have been increasingly becoming adept at industrial applications and interacting with human-designed spaces. Current capabilities extend to tactile exploration of unknown objects (e.g.,~\cite{regoli2017controlled,ohka2009object}), reasoning about complex manipulation objectives and motions (e.g.,~\cite{pairet2021path,ardon2021building}), and, among many other, human-robot and robot-robot cooperation (e.g.,~\cite{ardon2021affordance,feng2020overview}). Nonetheless, the possibility of adopting these advances in underwater applications seems to be far behind the ease of doing so in humanoid, terrestrial, and aerial robots. It is natural to wonder: \textit{what are the challenges that impede striving the potential of underwater manipulation to a fully autonomous level?} Understanding the limitations is key to the prospect of manipulators facilitating the strenuous, or even impossible, human labor in underwater recovery, intervention, and maintenance tasks.

\begin{figure}[t!]
    \centering
    \subfloat[ALIVE~\cite{evans2003autonomous} (2001-04)]{\includegraphics[width=0.9\columnwidth]{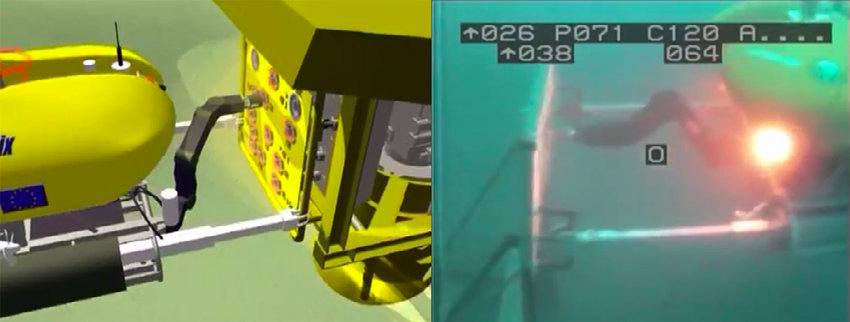}} \\
    \begin{tabular}{@{}m{\dimexpr0.475\linewidth-0.5em\relax}@{\quad}m{\dimexpr0.425\linewidth-0.5em\relax}@{}}
        \subfloat[TRIDENT~\cite{simetti2014floating,sanz2013trident} (2010-12)]{\includegraphics[width=\linewidth]{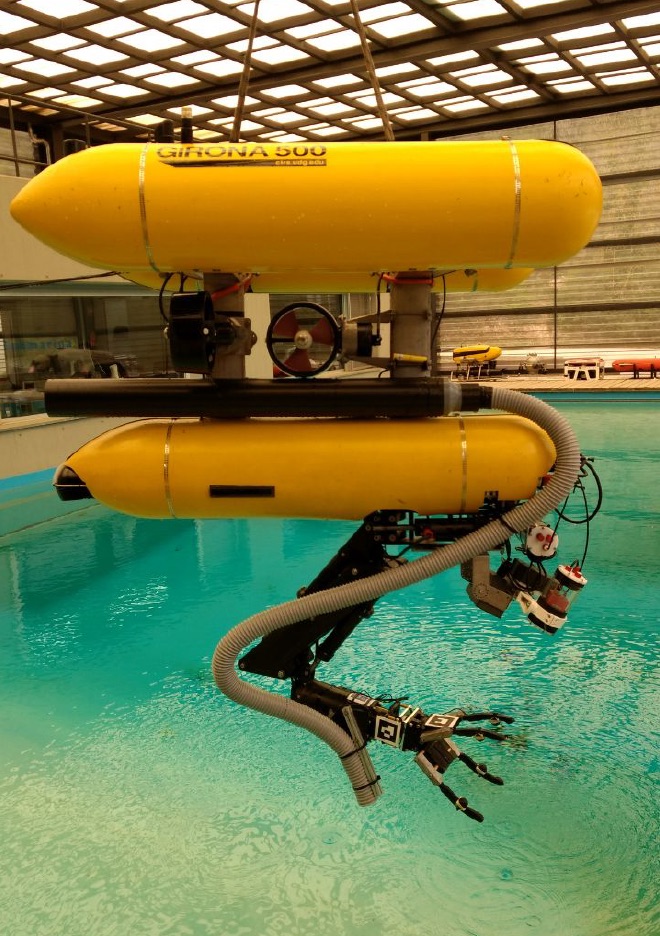}} &
        \begin{tabular}[b]{@{}c@{}}
            \subfloat[PANDORA~\cite{lane2012pandora} (2011-15)]{\includegraphics[width=\linewidth]{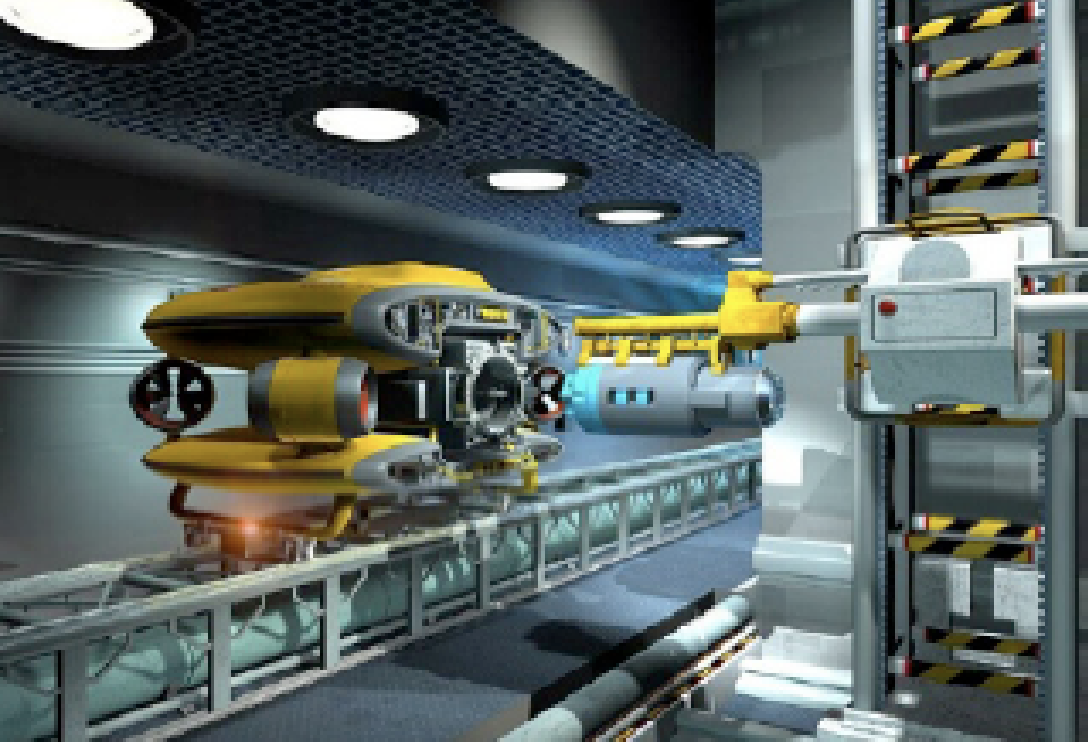}}\\
            \subfloat[OCEAN ONE~\cite{brantner2018controlling} (2016--)]{\includegraphics[width=\linewidth]{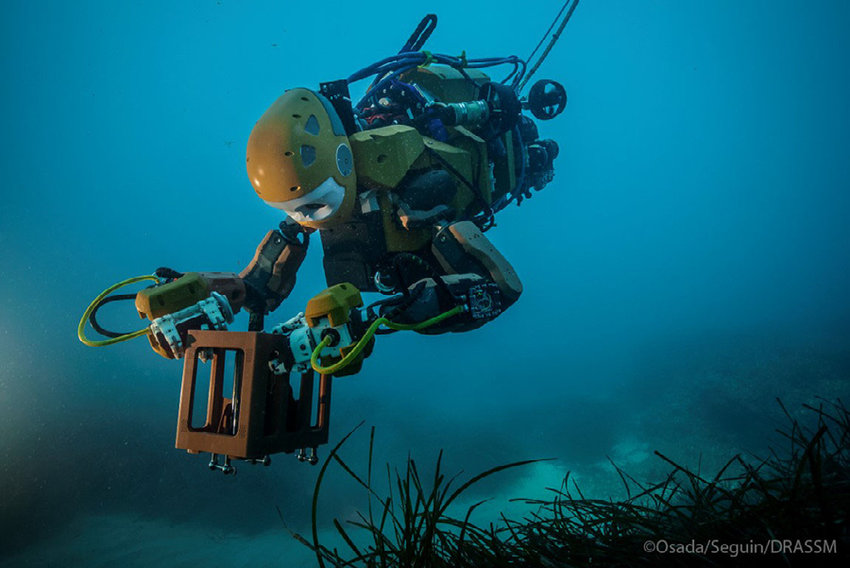}}
        \end{tabular}
    \end{tabular}
    \caption{Examples of underwater robot manipulators.}
    \label{fig:basis}
\end{figure}

The last decades have witnessed several notable contributions in the field of underwater manipulation, e.g.,~\cite{penalver2015visually,perez2015robotic,sivvcev2018underwater}.
Application-wise, most of the significant advances in the field have been achieved in multi-partnered research projects. Such works include: ALIVE~\cite{evans2003autonomous} (2001-2004) which was one of the first projects achieving a predefined valve-turning task, TRIDENT~\cite{simetti2014floating,sanz2013trident} (2010-2012) which performed teleoperated manipulation tasks, PANDORA~\cite{lane2012pandora} (2011-2015) that aimed at semi-autonomous manipulation, and most recently, OCEAN ONE~\cite{brantner2018controlling} (ongoing since 2016) which focuses on enhancing the operator experience during manipulation via \acl{HRI}.
Examples of the manipulators used in some of these projects are shown in \fref{fig:basis}. Although all these projects demonstrated progress towards autonomous underwater manipulation, their shared- and semi-autonomy methodologies fall short in comparison to other fields in robotics.

In this manuscript, we delve into the literature to extract the specific factors that make underwater manipulation such a challenging problem. The intended contribution of this manuscript, with respect to the state-of-the-art surveys, is to advance beyond a pure review. Instead, this paper identifies and discusses the limiting factors that impede the long-envisioned autonomy for underwater manipulation 
and correspondingly highlights promising avenues for future work.

\begin{figure*}
    \centering
    \resizebox{\textwidth}{!}{
    \begin{tikzpicture}
        \shade[rounded corners=0cm, top color=gray!20,bottom color=gray!80] (0,5) -- (0,5.80) -- (27.5,5.80) -- (28.5,5.45) -- (28.5,5.35) -- (27.5,5.00);
        
        \draw (0,5.40) node[circle,fill=secondary!45!white,draw]{1995};
        \draw (2,5.40) node[circle,fill=secondary!45!white,draw]{1997};
        \draw (4,5.40) node[circle,fill=secondary!45!white,draw]{1999};
        \draw (6,5.40) node[circle,fill=secondary!45!white,draw]{2001};
        \draw (8,5.40) node[circle,fill=secondary!45!white,draw]{2003};
        \draw (10,5.40) node[circle,fill=secondary!45!white,draw]{2005};
        \draw (12,5.40) node[circle,fill=secondary!45!white,draw]{2007};
        \draw (14,5.40) node[circle,fill=secondary!45!white,draw]{2009};
        \draw (16,5.40) node[circle,fill=secondary!45!white,draw]{2011};
        \draw (18,5.40) node[circle,fill=secondary!45!white,draw]{2013};
        \draw (20,5.40) node[circle,fill=secondary!45!white,draw]{2015};
        \draw (22,5.40) node[circle,fill=secondary!45!white,draw]{2017};
        \draw (24,5.40) node[circle,fill=secondary!45!white,draw]{2019};
        \draw (26,5.40) node[circle,fill=secondary!45!white,draw]{2021};
        \draw (27.8,5.40) node[circle,fill=secondary!45!white,draw]{2022};

        \filldraw[firstDBBar] (0,6.20) rectangle (5,6.40);
        \node[educationText] at (2.3,6.20){UNION 1993-99 \cite{rigaud1998union}};
        
        \filldraw[updatingBar] (0,4.40) rectangle (5,4.60);
        \node[educationText] at (2.3,4.30){AMADEUS 1993-99 \cite{lane1997amadeus}};
        
        \filldraw[updatingBar] (5,7) rectangle (7,7.20);
        \node[educationText] at (6,7){SWIMMER 1999-01 \cite{evans2001docking}};
        
        \filldraw[firstDBBar] (0,7) rectangle (2.5,7.20);
        \node[educationText] at (1.5,7){OTTER 1992-96 \cite{WangRL:1996}};
        
        \filldraw[primitiveBar] (17,7) rectangle (27,7.20);
        \node[educationText] at (22, 7)%
        {IEEE Technical Committee for Marine Robotics \cite{committee}};
        
        \filldraw[updatingBar] (7,4.40) rectangle (10,4.60);
        \node[educationText] at (8.2,4.30)%
        {ALIVE 2001-04 \cite{evans2003autonomous}};
        
        \filldraw[updatingBar] (3,3.60) rectangle (16,3.80);
        \node[educationText] at (8,3.60){SAUVIM 1997-09 \cite{marani2009underwater}};
        
        \filldraw[pyschologyBar] (17.5,3.60) rectangle (20,3.80);
        \node[educationText] at (18,3.60){TRITON 2012-14 \cite{triton}};
        
        \filldraw[pyschologyBar] (16,6.20) rectangle (19,6.40);
        \node[educationText] at (17.2,6.20){TRIDENT 2010-12 \cite{sanz2013trident}};
        
        \filldraw[pyschologyBar] (17,4.40) rectangle (21,4.60);
        \node[educationText] at (19,4.30){PANDORA 2011-15 \cite{lane2012pandora}};
        
        \filldraw[updatingBar] (22,6.20) rectangle (27,6.40);
        \node[educationText] at (24.3,6.20){OCEANS ONE 2016-- \cite{sanz2013trident}};
        
        \filldraw[firstWKBar] (25,3.60) rectangle (27,3.80);
        \node[educationText] at (26, 3.60)%
        {ICRA Workshop \cite{workshop1}};
        
        \filldraw[firstWKBar] (25,4.40) rectangle (27,4.60);
        \node[educationText] at (26, 4.30)%
        {EMRA Workshop \cite{workshop2}};
        
        \filldraw[pyschologyBar] (24,3.0) rectangle (27.5,3.20);
        \node[educationText] at (26, 3.0)%
        {TWINBOT \cite{pi2021twinbot}};

    \end{tikzpicture}
}
    \caption{Timeline of underwater projects that have achieved fully operated (orange), shared-autonomy (red) and fully-autonomous (blue) manipulation. Also, important gatherings (green) and other events (purple) that have had significant impact to the field.}
    \label{fig:timeline}
\end{figure*}
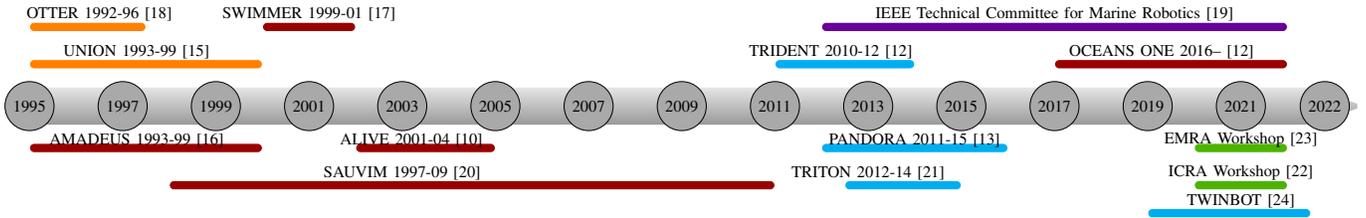

Our analysis starts in \sref{sec:synopsis} with a synopsis of previous research in the field of robotic underwater manipulation. We focus on result-oriented multi-partnered projects due to their unarguable relevant impact, as well as some key events that have promoted the progress of this line of research. Then, in \sref{sec:design_choices}, we extend this review by identifying critical design considerations that must be carefully examined when tackling underwater manipulation. Namely, we find to be of critical relevance the environmental conditions, hardware choices, both in the actuation and perception fronts, and the algorithmic stack.
The review on the design considerations motivates the discussion in \sref{sec:discussion} about the key challenges that must be addressed for the field to prosper. Finally, the paper concludes with final remarks in \sref{sec:final_remarks}.


    \section{SYNOPSIS OF UNDERWATER MANIPULATION \label{sec:synopsis}}



    This section provides a review on the development of \acp{I-AUV} since the early 90s. The progress of the field can be related to the \acp{I-AUV}' level of autonomy when performing a manipulation task. As such, we categorize the literature into three levels of intervention autonomy: (i)~fully teleoperated, (ii)~shared autonomy, and (iii)~fully autonomous. \fref{fig:timeline} illustrates a summary of the projects up to date that have performed underwater manipulation, altogether with important related events such as workshops and committees. 


    \subsection{Fully Teleoperated Manipulation}
    
        In this manuscript, we consider a fully teleoperated \ac{I-AUV} as that achieving a manipulation task by relying entirely on the teleoperators' skills for navigating and acquiring the target object. Different projects such as \ac{UNION}~\cite{rigaud1995ea} and \ac{OTTER}~\cite{WangRL:1996} have pioneered underwater manipulation by starting with fully teleoperated systems. For example, the \ac{ODIN} project proposed \ac{OBTLC} to relieve the user of the low-control management and allow the operator to focus on grabbing a target object~\cite{WangRL:1996}. The core outcomes of these early projects were tracking arbitrary objects underwater based on optical-flow and sign-correlation technology to obtain range and velocity maps from stereo cameras~\cite{WangRL:1996,rigaud1995ea}. In the \ac{OTTER} project, the authors demonstrated tracking of laser dot and arbitrary objects underwater for an object retrieval task. Besides semantic task planning, the automatic retrieval of objects requires stable low-control level task definition, but this means that a priori knowledge must be obtained \cite{otterdemo}.
        The current literature that heavily relies on teleoperation for underwater manipulation focuses on providing a stable control and navigation stack rather than on the target object detection and actions concatenation. All in all, the manipulation capabilities of fully-teleoperated systems are limited to the skills of the teleoperator, while being constrained by the need of low-latent communication.

    \subsection{Shared Autonomy Manipulation}
    
        Projects like \ac{AMADEUS}~\cite{lane1997amadeus} and \ac{ODIN}~\cite{choi1995development} propose teleoperated-based approaches where the system obtains feedback from the end-effector to increase the changes of grabbing an object. For example, one of \ac{AMADEUS}'s main goals was to improve dexterity by integrating the three-fingered gripper with force and slip contact sensing. At a later stage, \ac{AMADEUS} demonstrated coordinated motion of two manipulators on a rigid object inside a water tank. In \ac{ODIN} project, the team designed and developed their vehicle to cope with dual operational modes, shared autonomy, and tethered, to operate a single \ac{DoF}, mechanical manipulator, with force open and closure feedback~\cite{choi1995development}. While the works in this section receive sensing feedback from the end-effectors, they rely on a teleoperator indicating the suitable grasping and action plans to interact with an object.

    \subsection{Full Autonomous Manipulation}
    
        The works in this section perform manipulation underwater fully autonomously. This means they are able to detect an object, plan grasps and actions on it without the intervention of a user. Some of the projects that have demonstrated such autonomous manipulation capabilities are \ac{SAUVIM}~\cite{marani2009underwater}, TRIDENT~\cite{sanz2013trident}, and \ac{TWINBOT}~\cite{pi2021twinbot}.
     
        In 2005, \ac{SAUVIM} underwent experiments to validate its manipulation capabilities, achieving hooking a cable on a target, searching for a cable and cutting it. To accomplish these tasks, \ac{SAUVIM} used extensive sensor data processing through a multiplexer that allowed the system to discretize sensory input given distances to the target thresholds. Other projects such as TRIDENT~\cite{sanz2013trident} and \ac{TWINBOT}~\cite{pi2021twinbot} take advantage of visual sensory input to detect target objects, mainly using machine-readable optical labels. For example, \ac{TWINBOT} demonstrated grasping and high-accuracy cooperative transportation of a pipe from one stand to another in a priori optical map. The two \acp{AUV} followed a predefined waypoints sequence to accomplish this task which resulted in the effective placement of the pipe on the next stand. While the most common technique is the use of machine-readable codes for object detection, there are some steps towards object segmentation by using specialized laser scanners~\cite{pi2021twinbot} that allow the system to discern the target object. A decentralized \ac{TP} kinematic control algorithm was also implemented in the \ac{TWINBOT} project to tackle the issue of highly limited communication bandwidth underwater. \ac{TWINBOT} experimentally validated modularity and robustness with the use of grouped modes to allow changes in the system's behavior \cite{pi2021twinbot}. 
        
        Moreover, although some advancements have been made towards autonomous manipulation,  there remain many avenues for improvement compared to robotic manipulation state-of-the-art in dry environments. A discussion on possible future research paths can be found in \sref{sec:discussion}.



    \section{DESIGN CONSIDERATIONS \label{sec:design_choices}}

\begin{figure}[b!]
        \centering
\hypersetup{linkcolor=black}

\resizebox{\columnwidth}{!}{
    \begin{tikzpicture}
    	\tikzstyle{every node}=[font=\LARGE]
        \path[mindmap,concept color=gray!80!white,text=black]
        node[concept,minimum size=4.2cm] {\textbf{Design Considerations}}
        [clockwise from=-10]                    
        
        child[concept color=gray!65!white, level distance=4.2cm,concept color=orange!85!black, style={sibling angle=0}] {
          node[concept,minimum size=3cm] {\!Arm Hardware}
          [clockwise from=90]
          child [level distance=3cm] { node[concept,minimum size=2.5cm] {\acs{DoF}} }
          child [level distance=3cm] { node[concept,minimum size=2.5cm] {\!\!\!\!Actuators} }
          child [level distance=3.1cm] { node[concept,minimum size=2.5cm] {\!\!\!Gripper type} }
          child [level distance=3cm] { node[concept,minimum size=2.5cm] {\!\!\!Finger sensor} }
        }    
         child[concept color=gray!65!white, level distance=4.2cm,concept color= blue!45!white, style={sibling angle=-110}] {
          node[concept,minimum size=3cm] {Sensory \& Data Collection}
          [clockwise from=150]
          child[level distance=3.2cm] { node[concept,minimum size=2.5cm] {\!\!visual}}
          child[level distance=3.2cm] { node[concept,minimum size=2.6cm] {\!\!\!sonar}}
          child[level distance=3.2cm] { node[concept,minimum size=2.5cm] {\!\!laser}}
          }
        child[concept color=gray!65!white, level distance=4.4cm,concept color=cyan, style={sibling angle=-100}] {
          node[concept,minimum size=3cm] {Algorithmic Control}
          [clockwise from=230]
          child [level distance=3cm] { node[concept,minimum size=2.3cm]  {\!\!Motion} }
    	  child [level distance=3cm] { node[concept,minimum size=2.3cm] {Force} }
    	  child [level distance=3cm] { node[concept,minimum size=2.3cm] {Data Management} }
        }
          child[concept color=gray!65!white, level distance=4.5cm,concept color=green!75!black, style={sibling angle=-90}] {
          node[concept,minimum size=3cm] {\!\!Environ-ment}
          [clockwise from=-30]
          child [level distance=3.2cm] { node[concept,minimum size=2.3cm]  {\!\!\!Distur-bances} }
          child [level distance=3.2cm] { node[concept,minimum size=2.3cm]  {\!\!\!Visibility} }
          child [level distance=3cm] { node[concept,minimum size=2.3cm]  {Depth} }
        }
        ;
    \end{tikzpicture}
}
        \caption{Design choices to consider when aiming at autonomous underwater robot manipulation. We outline the options found in the literature in regards to hardware (orange), sensory input (purple), control (blue) and environmental conditions (green). We build onto these challenges to motivate future work that can thrive the potential of underwater robot manipulation.}
        \label{fig:mindmap}
    \end{figure}
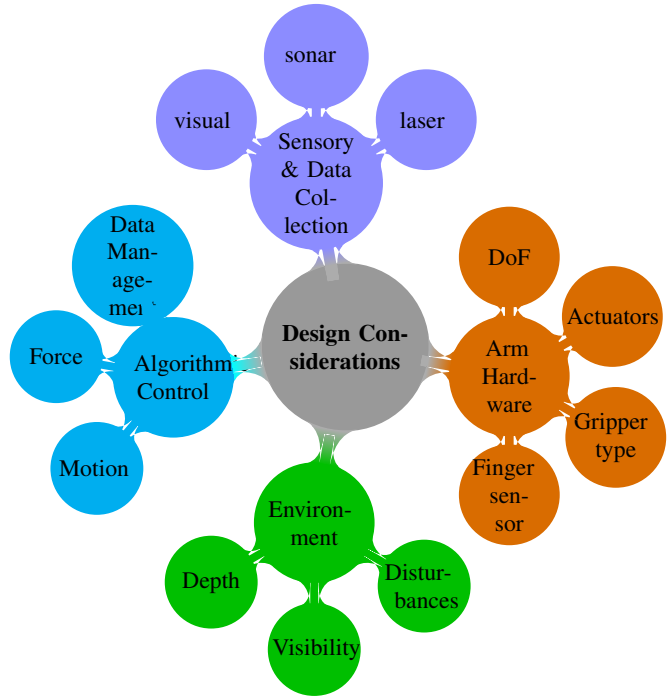

    After reviewing the autonomous capabilities of past projects in \sref{sec:synopsis}, we 
    hypothesize there are common design considerations to the design and implementation of underwater robotic manipulation. As illustrated in \fref{fig:mindmap}, we extract four design considerations: the environmental conditions where the \ac{I-AUV} has to perform, leading to the choice of sensory input and arm hardware, as well as algorithmic stack for the control. Moreover, other considerations such as the \ac{I-AUV}'s payload capacity play an important role when designing manipulation techniques. The payload aspect is closely related to control stability and is extensively reviewed in~\cite{sivvcev2018underwater}. 
    
    \subsection{Environment Dynamics}
    
        \subsubsection{Disturbances} 
            
            
            Hydrodynamic forces raising from buoyancy, lift, drag, and, among many others, added mass have become more feasible to quantify over the last decades~\cite{fossen2011handbook}. However, forces originated by external factors such as currents and waves are more challenging to predict. As thus, these disturbances heavily affect subsea vehicles, specially those that require high-operating precision such as \acp{I-AUV}. The impact of underwater disturbances to the co-dependent vehicle-arm system requires manipulation strategies that consider the system as a whole; the stronger the disturbances, the more robust the control of the whole vehicle needs to be.
        
        \subsubsection{Visibility} 
           One of the main challenges for vision-dependent subsea experiments is poor visibility which affects intervention missions (i.e. detecting a valve for grasping and turning, identifying a connector for a plugging/unplugging task, or object identification for recovery, among others). Common environmental causes of reduced visibility are water turbidity, sun reflections, and changes in depth. In turn, these environment conditions lead to poor \ac{2D} video feedback and compromised \ac{3D} perception \cite{yu,penalver2015visually,SIVCEV2018153}. The negative effects on sensory input are notorious in autonomous manipulation missions, given that current algorithms do not cope with uncertainty in the input data. Generally, the low quality of visual feedback also affects skilled professionals engaged in teleoperation missions~\cite{sivvcev2018collision}.
           
        \subsubsection{Depth} 
            The deeper the intervention mission, the more robust the \ac{I-AUV} and its algorithmic stack need to be. Depth not only affects the vehicle pressure and sensory feedback but also challenges the feasibility of aiding the manipulation via teleoperation. During ocean trials, depth rating of manipulator arms becomes an even more important factor. Many commercial manipulator arms offer a depth rating of at least 300 m which is suitable for most manipulation applications~\cite{sivvcev2018underwater}.
        
    \subsection{Arm Hardware}
            Differently from other robotic fields and given the difficult nature of the marine settings, underwater robotic arms are not as technologically advanced as manipulators in dry setups. For example, despite the well known singularity problem and constraints involving less than 6 \ac{DoF} manipulators, most of the arms for underwater applications range between 4 and 5 \ac{DoF} \cite{sivvcev2018underwater}. Most of the current literature relies on the \ac{DoF} provided by the mobile base, although falling into other algorithmic problems such as stable control and planning for mobile base manipulation. Furthermore, in dry settings, robotics technologies are starting to move towards dexterous hands. However, underwater end-effectors are still limited to mostly 2 finger grippers \cite{yuh2001underwater}. Additional to grippers, actuators and available \ac{DoF} determine the arm hardware specifications. 
            
        \subsubsection{Actuators}
            Actuators for underwater manipulators can roughly be divided into hydraulic or electric. Hydraulic actuators present do not require mechanical components (e.g., gears and levers) to eject force, thus offering a more efficient power to weight ratio~\cite{gary2004water,yao2009development,zhang20147000m}. On the other hand, electric actuated manipulators, similar to industrial manipulators, although heavier, provide a more precise motion and force/torque control~\cite{sivvcev2018underwater,ishitsuka2007development,yoerger1991design}. There are other modalities such as pneumatic and tendon-driven actuators that are popular for underwater soft-robotic applications. A detailed review of such actuation methodologies is detailed in~\cite{wang}.
            
        \subsubsection{\ac{DoF}} 
            
            Depending on the manipulation task, it is usually preferred to have a redundant manipulator.
            Redundancies ease reachability and, at some level, autonomy in the case of obstacle avoidance for manipulation purposes~\cite{marani2009underwater,ribas2015auv}. However, up to date, it is common that most of the underwater manipulation missions fall into objects retrieval. In some cases, the manipulation relies on the \acp{DoF} provided by the mobile base to achieve the mission~\cite{sivvcev2018underwater,spong2020robot}.
        
        \subsubsection{Type of gripper} 
           
           Nowadays, underwater manipulators come with interchangeable grippers which purpose range from grabbing different objects to cutting an abandoned underwater net, for example. The most common gripper type is the two-parallel finger gripper \cite{blueprint,sivvcev2018collision}. 
           Most common actuators used to eject the gripping force are hydraulic, but more recently, electrical actuators have been used for this purpose~\cite{sivvcev2018underwater}.

        \subsubsection{Finger sensors} 
          For underwater manipulation, replicating tactile sensing has proven to be a hard task. The most commonly used finger sensors are based on waterproof actuators that drive the fingers by means of a cable based transmission~\cite{blueprint}. Force/torque sensors are typically integrated on each for grasp and force control \cite{ribas2015auv,scarcia2015design}.
        
    \subsection{Sensory Input } 
        The sensory input is related to the data acquisition. It refers to the medium used to recognize all the physical and visual qualities that suggest a manipulation action in the scene. Differently from conventional dry robotic manipulation, in underwater settings, sensory input is affected by limited accessibility and the inherent dynamics of the environment. These environment changes result in flickering, insufficient illumination, refraction of the sensory input, and more. Next, we summarize the most commonly used type of sensors employed to detect target objects for underwater manipulation.
        
        \subsubsection{Visual input} 
            Underwater imaging technologies have shown a fast progress in the last decade, as summarized in~\cite{cong2019novel}. Most of the underwater visual sensors can achieve high resolution and accuracy when in short distance to the target~\cite{cong2019novel,hong2019water,akkaynak2019sea}. This property is specially relevant for manipulation tasks such as detecting details in archaeological applications~\cite{akkaynak2019sea}, object tracking~\cite{cong2019novel}, and retrieval~\cite{hong2019water}. 
     
        \subsubsection{Sonar input} 
            Visual sensory input is highly constrained by the limited light transmission in underwater environments. Thus, sonar sensing is a widely adopted alternative~\cite{yu2006development,kallasi,jonsson2009observing,gordon1992use}. Sonar is specially used for scene reconstruction and localization. For example, \cite{yu2006development} uses a \ac{3D} sonar imaging system for object recognition based on sonar array cameras. Although the sonar imaging started with lower resolution than purely visual input, there has been progress towards high-resolution sonar cameras \cite{yu2006development}. Nonetheless, there are still issues related to blurred edges and reflections that makes this type of sensory input still not ideal for underwater manipulation.
            
        \subsubsection{Laser light input} 
            Although expensive, \ac{3D} laser scanners have shown to be an alternative to \ac{3D} cameras for the underwater environment \cite{chi2016laser,liu2010practical,palomer20183d}. \cite{palomer20183d} proposes an in-house designed casing that, differently from stereo camera imaging, can operate up to 5~m distance. Specifically for underwater manipulation, the experiment in \cite{palomer20183d} focused on picking up an amphora from the bottom of a water tank, without considering reliable grasping configurations nor obstacle avoidance. Nonetheless, \cite{palomer20183d} achieves an exceptional detection and single object segmentation performance.

    \subsection{Control}  
    
        Subsea intervention tasks carried out with underwater manipulators often demand extensive contact with the environment. Therefore, manipulator trajectory tracking control has to be backed up with the implementation of interaction force-torque control with the environment. Methodologies should consider that slight deviations of the end-effector from the planned trajectory can cause the manipulator to either lose contact with the surface or generate large interaction forces that lead to tear~\cite{spong2020robot}.

        \subsubsection{Force}
         
        One of the first works addressing the force interaction problem for an underwater manipulator was~\cite{dunnigan1996hybrid} who propose a hybrid position and force control scheme. Additional research on underwater manipulator force control algorithms with validation through extensive numerical simulations can be found in~\cite{kajita1997force,cataldi2015basic}.

    \subsubsection{Motion} 
    
        \acp{I-AUV} tested thus far suffered from  the environmental disturbances and dynamic coupling, as previously mentioned, and  research  on  the  control  schemes for  underwater  manipulators  only  extend  to  simulations. \acp{I-AUV} must be able to adjust to these variations in the environment in order to yield successful results. Attempts have been made to stabilize the system for example having adaptive controls. Earlier iterations of \acp{I-AUV} were testbeds for understanding this particular challenge such as in the cases of \ac{ODIN} and \ac{UNION}. \ac{ODIN}'s control schematics utilizes \ac{PD}, neural network controllers, and an adaptive controller which is used for parametric deviations due to vehicle dynamics\cite{choi1995development}. Stabilization of the system also extends to navigation with different geomagnetic, optical, bathymetric algorithms to aid in a more steady navigation \cite{gonzalezgarcia}. 
        The last two decades have nevertheless witnessed an  increasing number of proposals to circumvent this inevitable control issue. Underwater manipulators have generally been designed ad-hoc as their commercial counterparts are expensive and do not allow customizing with their integrated motion controller requiring more changes to the whole system. Further development of an adaptive control scheme promises more reliable intervention results.
        
    \subsubsection{Data Collection}
        The data acquisition is closely related to the sensory input. Together with the sensors used for collecting data, one usually also decides on the underlying data structure. Depending on the sensory used, the collected data can be limited to images, or it can be a combination of visual-and-sonar or visual-and-laser data structures. Methods based on acoustic and optical fusion reconstruction combine the advantages of both sensory inputs and improve underwater performance. Examples of such methodologies are presented by \cite{galcerancoverage,kucukkaya2004photogrammetry,massot2015optical}. Although these works' contributions are solely for image reconstruction, their findings have potential to be used for underwater manipulation.

    \section{DISCUSSION and FUTURE\\RESEARCH AVENUES \label{sec:discussion}}

    As a summary, we classified the works according to their level of autonomy in \sref{sec:synopsis}, starting from fully teleoperated, shared-autonomy, and fully autonomous manipulation underwater. Next, \sref{sec:design_choices} details the different design considerations made in the development of underwater manipulation systems. On the strengths of the current literature, we can highlight the technological advancements on light and sonar sensory that allow to have a more accurate perception and potential object reconstruction. Despite such progress, the state of the manipulation underwater does not compare to the autonomy levels that have been reached on robotic manipulation in dry or air environments.
    
    For example, when it comes to object perception, underwater manipulation faces visibility issues besides the potential occlusions produced by the arm. Potential dirt and changes in light make it difficult to learn a model that detects and recognize possible target objects for manipulation. Moreover, water-glass-air interfaces between the sensor and the scene modify the intrinsic parameters of the visual sensors and limit the performance of the image processing algorithms. Currently, the literature overcomes this issue with a combination of sonar sensors and cameras or by using computer-readable codes. Recent active \ac{3D} marker designs reveal promising results for visibility improvement yet still require further development for highly-accurate pose estimation~\cite{visibility}. In this regard, state-of-the-art systems require knowledge of the object and its pose in advance. As a result, current underwater manipulation does not consider online grasp planners on familiar or novel objects as part of their manipulation pipeline. Along the same lines, there are no public datasets dedicated to underwater robotic manipulation. Other limitations related to hardware and motion stability are further addressed in \cite{sivvcev2018underwater}.

    
    In general, \ac{I-AUV} technology has progressed considerably since the early 90s, but there is still a long way to go before full autonomy intervention is achieved. Focus on improvements with underwater communication, sensing, data acquisition, and data processing will be the next steps towards increasing levels of autonomy. In the meantime, to overcome these obstacles, state-of-the-art in \acp{I-AUV} recovery missions have divided interventions missions into multiple phases. For example, for an object recovery mission, there would be the mapping scene for localization and object detection phases. For the mapping scene phase, is responsible for surveying the location and collecting images typically using a down-looking camera to achieve photo mosaicking. In another phase, the \ac{I-AUV} explores the area and identifies objects for recovery. Given the dynamic nature of the underwater environment, multiple-phased missions pose extra challenges. Surveying and identifying a region of interest takes a few hours depending on the area, during which the region may be disturbed by natural occurrences like currents or fauna activities, as well as \ac{I-AUV} disturbances (if too close to the site). Once deployed for the intervention phase of the mission, the area may have changed slightly. This change can cause a bigger challenge, given that the \ac{I-AUV} will be returning with processed data to utilize for the mission. Thus, ideally, \acp{I-AUV} should be deployed in one single stage where surveillance and intervention are done as part of the same phase. This method can reduce the duration of the mission and the need for more iterations over the area. Moreover, there are visibility issues. Submerged objects over time become veiled with sand, flora, fauna, rusting and require an increasing degree of intervention to uncover in order to recognize them. Given that object detection and recognition are highly dependent on features like color, texture, contours, and intensity, data collection for computer vision underwater still suffers from image intensity degradation, haze effect, color distortion requiring several pre-processing and post-processing techniques before effectively extracting features. Up to date, feature extractors like oriented FAST and rotated BRIEF (ORB) have been commonly used \cite{RIDAO2015227}. The further study of customized feature extractors can potentiate real-time applicability on object recognition for manipulation purposes.

    The authors also recognize an ongoing challenge regarding long-term navigation and multi-agent task planning that are closely related to stable autonomous manipulation. Overcoming this feat will open opportunities in areas still in their preliminary stages of development. Namely for cooperative missions, using a robot-robot approach when performing manipulation tasks might entail a drop in the mission time and improve navigation and mapping quality. Moreover, a collaborative setting will allow the system to perform in more complex scenarios, such as autonomous valve turning, \ac{SAR}, and hot stab operations. The main issue encountered in collaborative subsea robotics is the lack of an underwater communication standard. 
    Moreover, greater efforts should be invested in \ac{TP} algorithms for single \ac{I-AUV} and collaborative tasks. Up to date, \ac{TP} algorithms have been a recent solution to handle multiple sequential tasks such as maintaining target object visibility, end-effector configuration, and handling manipulator joint limits \cite{adaptiveadmittance,simetti2014floating}. \ac{TP} has been an efficient approach, specially for coordinated control of \acp{I-AUV} to perform autonomously \cite{adaptiveadmittance}. Nonetheless, \ac{TP} for kinematic control remains a topic requiring much improvement to progress towards autonomous manipulation missions \cite{simetti2018}.

    In terms of challenges with regards to sensing technology, there remains room for advancements towards a combination of various sensing modalities. Sonar and visual sensor fusion has been integrated in recent experiments to produce point cloud datasets~\cite{gonzalezgarcia,nico,shaukat} with the goal of improving results in localization, communication, and object recognition. To this date, this sensory combination has enhanced the \ac{I-AUV} state estimation, thus allowing for better performance~\cite{nico}. Further experimental validation with sensor fusion, such as \ac{INS} measurements integrated with acoustic and vision-based systems, will allow for higher accuracy in positioning for short ranges \cite{kanda}.

    \section{FINAL REMARKS \label{sec:final_remarks}}
    

    In this survey, we explored the literature for approaches that focus on underwater manipulation and identified different levels of autonomy when performing the manipulation task. In contrast to previous reviews for underwater manipulators, we discussed the environmental, hardware, and algorithmic factors that challenge autonomous underwater manipulation. In the light of the identified limitations, we highlighted several problems in the field and future research avenues to potentiate underwater robotic manipulation.

	\bibliographystyle{ieeetr}
    \bibliography{references}
\end{document}